\documentclass[conference]{IEEEtran}

\usepackage{cite}
\usepackage{graphicx}
\usepackage{booktabs}
\usepackage{array}
\usepackage{url}
\usepackage{balance}
\usepackage{amsmath}
\usepackage{amssymb}

\title{Machine-Learning Assessment of the Predictive Value of Inflammatory Biomarkers for Cognitive Impairment in an Older Hispanic Adult Cohort}

\author{
\IEEEauthorblockN{\begin{tabular}{c}
Antony Garcia\textsuperscript{1,3}, Gabrielle Britton\textsuperscript{2}, Alcibiades E. Villarreal\textsuperscript{4}\\
Diana C. Oviedo\textsuperscript{4,5,6}, Giselle A. Rangel\textsuperscript{4,5}, Xinming Huang\textsuperscript{1}
\end{tabular}}
\IEEEauthorblockA{\textsuperscript{1}Worcester Polytechnic Institute, Worcester, MA, USA\\
\textsuperscript{2}Centro de Vacunaci\'on e Investigaci\'on (CEVAXIN), Panam\'a, Panam\'a\\
\textsuperscript{3}Facultad de Ingenier\'ia El\'ectrica, Universidad Tecnol\'ogica de Panam\'a, Panam\'a, Panam\'a\\
\textsuperscript{4}Instituto de Investigaciones Cient\'ificas y Servicios de Alta Tecnolog\'ia (INDICASAT AIP), Panam\'a, Panam\'a\\
\textsuperscript{5}Sistema Nacional de Investigaci\'on (SNI), Secretar\'ia Nacional de Ciencia,\\
Tecnolog\'ia e Innovaci\'on (SENACYT), Ciudad del Saber, Panam\'a\\
\textsuperscript{6}Escuela de Psicolog\'ia, Universidad Cat\'olica Santa Mar\'ia La Antigua, Ciudad de Panam\'a, Panam\'a}
}

\begin{document}
\maketitle

\begin{abstract}
Small clinical tabular datasets need interpretable machine learning because deep learning is not viable and ensembles are hard to inspect. A pitfall is that statistical significance is not predictive utility. Using data from the Panama Aging Research Initiative--Health Disparities (PARI-HD) cohort ($n=165$), a leakage-safe threshold-likelihood Bernoulli/Categorical Naive Bayes (BNB/CNB) classifier was implemented: each continuous predictor was reduced, inside every training fold, to a supervised chi-square state; income entered as a categorical likelihood. Every data-dependent step occurred inside repeated stratified ten-fold cross-validation (30 repeats). The demographic baseline reached ROC-AUC $0.630\pm0.017$. I-309 (CCL1) was the dominant incremental feature ($\Delta$AUC $+0.110$; paired DeLong below $0.05$ in $100\%$ of repeats) and, as a pre-specified primary analysis, gave a fixed-partition DeLong $p=0.0018$ validated across 200 random partitions (median $p=0.0011$). Under the 18-candidate exploratory family, I-309 reached Benjamini-Hochberg $q=0.032$ on the frozen partition and cleared $q<0.05$ in $85\%$ of partitions, while no other marker showed reliable incremental value. Because the fitted model is an inspectable table of thresholds and probabilities, I-309/CCL1 is presented as an interpretable candidate feature for tabular prediction of cognitive impairment, pending external validation.
\end{abstract}


\section{Introduction}

Infection and chronic inflammation are implicated in neurodegeneration and cognitive decline. Rangel et al. reported that cumulative pathogen exposure, \emph{Chlamydia pneumoniae} seropositivity, and several serum inflammatory markers were associated with cognitive impairment in older Hispanic adults from the PARI-HD cohort \cite{rangel2025}. That study was an association analysis adjusted for age, sex, education, and income. A question remains: when the same data are treated as a prediction problem, is any statistically associated biomarker also useful as an individual-level predictive feature beyond those covariates?

In a small cohort, this question is difficult. A biomarker can appear useful under one train--test split yet lose value under another, so a useful feature must remain informative across repeated partitions. Flexible models overfit; preprocessing such as thresholding or imputation can leak test information into training and inflate performance; and screening markers manufactures nominal ``hits'' unless multiplicity is controlled. Without these safeguards, biomarker-panel predictive value can be overstated.

This study presents a reproducible predictive analysis of the PARI-HD panel that separates statistical association from machine-learning feature utility. First, a threshold-likelihood BNB/CNB framework \cite{thresholdbnb} is used as an auditable model artifact, a table of class priors and conditional probabilities, rather than as a new method. Second, every data-dependent operation, including threshold selection, categorical likelihood estimation, and recalibration, is performed inside cross-validation folds, so reported performance is leakage-safe. Third, a single primary hypothesis (I-309/CCL1) is separated from the exploratory panel, and sensitivity to the cross-validation partition is quantified. The goal is not a deployable diagnostic but an answer to a question: among biomarkers found relevant by association analysis\cite{rangel2025}, is any single marker also useful for incremental prediction beyond demographics, and how stable is that answer? All results presented in this manuscript can be reproduced from the released code and aggregate materials \cite{garcia2026repo}.

\section{Related Work}

The PARI-HD cohort was introduced as a study of older adults in Panama recruited through the Geriatrics Service of the Dr. Arnulfo Arias Madrid Hospital Complex and used by Rangel et al. to examine infectious exposure, inflammatory biomarkers, and cognitive impairment \cite{rangel2025}. In that study, serum IgG reactivity to seven pathogens was measured by ELISA, and serum inflammatory proteins were quantified with a multiplex Simoa platform. Descriptive analyses were followed by univariate tests, including ANOVA for continuous variables and Pearson $\chi^2$ tests for categorical variables. Multivariable logistic or linear regression models used age, sex, education, and income as covariates; doubtful positives were reclassified as negative, missing data were excluded, and cumulative infectious exposure was the summed number of IgG-reactive pathogens \cite{rangel2025}.

In that analysis, infection and inflammation signals were associated with cognitive diagnosis or cognitive-functional measures. \emph{Chlamydia pneumoniae} seropositivity differed between cognitively unimpaired and impaired groups ($p=0.02$), and TNF-$\alpha$ was associated with \emph{C. pneumoniae} seropositivity (OR $=2.08$, 95\% CI $1.0$--$4.1$, $p=0.04$). Mean I-309 and TNF-$\alpha$ levels were higher in cognitively impaired participants ($p<0.001$ and $p=0.012$). Each additional pathogen exposure was associated with greater odds of cognitive impairment (OR $=1.51$, 95\% CI $1.01$--$2.26$, $p=0.04$), and poorer Trail Making Test A performance was associated with cumulative exposure (OR $=17.43$, 95\% CI $2.32$--$32.53$, $p=0.02$). These findings create a prediction question: do I-309, TNF-$\alpha$, or the inflammatory panel add discrimination once the demographic baseline is used?

These findings are consistent with literature linking infection, inflammatory burden, and chemokine activity to AD and mild cognitive impairment (MCI). \emph{C. pneumoniae} has been associated with AD in meta-analytic work on bacterial infection \cite{maheshwari2015}, and infectious burden including \emph{C. pneumoniae} has been associated with AD and cognitive performance \cite{bu2015}. CCL1/I-309 is biologically plausible: chemokine alterations through the AD--MCI spectrum, including CCL1/I-309 signals, were found in a systematic review and meta-analysis of blood and cerebrospinal-fluid chemokines \cite{zhou2023}, and altered blood proteins tracking AD progression were reported in longitudinal proteomic work from the Australian Imaging, Biomarkers and Lifestyle (AIBL) study \cite{gupta2017}. In elderly Hispanic populations, AD and MCI have been distinguished from normal cognition by serum protein profiles \cite{villarreal2016}. TNF-$\alpha$ is supported by reports of increased TNF-$\alpha$ in subjective cognitive impairment \cite{serafini2024} and by studies connecting infection burden with inflammatory cytokines in AD \cite{bu2015}.

Naive Bayes classifiers remain attractive for this association-to-prediction translation because the fitted model reduces to class priors and per-feature conditional probability tables that a clinician can read; the conditional independence assumption is well characterised \cite{domingos1997}. This choice suits the data regime: with a small cohort, deep learning is not viable, and while flexible ensembles can be competitive they are not directly interpretable. Rudin argued that, for high-stakes decisions, interpretable models should be preferred over post-hoc explanations of black boxes \cite{rudin2019}. That philosophy is followed by the threshold-likelihood BNB/CNB framework used here \cite{thresholdbnb}: continuous predictors are binarised through a supervised threshold, an idea with a history in supervised discretization \cite{fayyad1993}, while categorical predictors are retained and missing terms are omitted from the likelihood.

For prediction, discrimination is only half of model quality; calibration, the agreement between predicted risk and observed frequency, is frequently neglected yet decisive for threshold-based decisions \cite{vancalster2019}. For multiplicity, the false-discovery rate is controlled by the Benjamini-Hochberg procedure \cite{benjamini1995}, while privileging pre-specified primary hypotheses before an exploratory family is formalised by gatekeeping designs \cite{dmitrienko2011}. Paired comparisons of correlated ROC curves are used throughout \cite{delong1988}, and beta calibration is used as a leakage-safe recalibration check \cite{kull2017}. These ingredients are combined into a reproducible workflow in which both a stable primary predictive feature and an underpowered panel-wide screen are reported.

\section{Methods}

\subsection{Cohort and Outcome}
The dataset contained 165 PARI-HD participants aged at least 65 years. The binary outcome contrasted cognitively unimpaired participants ($n=69$) with cognitively impaired participants (mild cognitive impairment, Alzheimer's disease, vascular dementia, or mixed dementia; $n=96$). The sample had mean age $80.0\pm7.7$ years, 110 women ($66.7\%$), and mean education $7.6\pm3.9$ years. Diagnosis-defining or assessment variables were excluded from every model, including cognitive test scores, functional scores, diagnosis codes, memory complaints, and the case--control flag; consequently no MMSE, GDS, clock-drawing, trail-making, or ADL variable was available to the classifier.

\subsection{Baseline and Candidate Family}
To make the prediction question parallel to referenced study, the baseline was matched to the previously evaluated covariates: age, sex, education, and income \cite{rangel2025}. Age and education were treated as continuous thresholded BNB terms, and sex as a binary term. Income was represented as a CNB as four source-study categories ($<250$, $250$--$500$, $501$--$850$, $>850$ U.S. dollars).

The discovery family was frozen before evaluation to ask whether variables studied in the source association framework behaved as predictive features: 18 inflammatory markers referenced in \cite{rangel2025}.

\subsection{Threshold-Likelihood Model}
For each continuous predictor, candidate thresholds were evaluated within the training fold, following the threshold-likelihood binarization framework of \cite{thresholdbnb}. The selected threshold $t^\ast$ maximised the $\chi^2$ statistic of association between the binarised feature (state $x \ge t$) and the class label, subject to a minimum leaf size of 3. This $\chi^2$ criterion is supervised binarization: it is estimated inside the training fold and applied to held-out data, so it selects a cut point rather than serving as an inferential test, and no per-threshold correction is applied. Multiplicity across the biomarker family is controlled by the BH-FDR procedure described below. The Bernoulli likelihood was used for binary variables without thresholding, and a categorical likelihood over four levels for income. At prediction time, missing values were omitted from the log-likelihood sum rather than imputed, so each patient was scored on observed markers. The fitted artifact, including priors, thresholds, and probability tables, is inspectable.

\subsection{Cross-Validation and Incremental Test}
Evaluation used repeated stratified ten-fold cross-validation with 30 repeats. The repeat count was a stability check: one ten-fold partition leaves about 16--17 participants per test fold, so repeated partitions reduce dependence on any single split and identify features that stay useful across train--test assignments, without treating the repeats as independent studies. Within each repeat, matched out-of-fold (OOF) scores were produced for every participant by the baseline and each baseline-plus-candidate model, and ROC-AUC was computed from those scores. For a candidate feature $f$, let $\mathrm{AUC}_0$ and $\mathrm{AUC}_f$ denote the OOF ROC-AUC of the baseline and the baseline-plus-$f$ model on the same participants. The paired DeLong test \cite{delong1988} asks whether $f$ adds incremental discrimination:
\[
H_0:\mathrm{AUC}_f=\mathrm{AUC}_0
\qquad\text{versus}\qquad
H_1:\mathrm{AUC}_f\neq\mathrm{AUC}_0.
\]
Under $H_0$, the feature adds nothing beyond the baseline. The two-sided statistic was
\[
z=\frac{\mathrm{AUC}_f-\mathrm{AUC}_0}{\widehat{\mathrm{SE}}(\mathrm{AUC}_f-\mathrm{AUC}_0)}
\qquad\text{with}\qquad
p=2\Phi(-|z|).
\]
For each candidate, mean AUC, mean $\Delta$AUC versus baseline, the median two-sided paired DeLong $p$-value over repeats, and the percentage of repeats with $p<0.05$ are reported. Each biomarker was thus judged by incremental discrimination rather than by association alone. The percentage of significant repeats is a resampling-stability descriptor only; it is not an independent hypothesis test and is not pooled.

For the formal discovery test, one paired DeLong $p$-value per candidate was generated from a pre-specified ten-fold partition (seed). Because a single partition is fragile at this sample size, the primary feature's $p$ was recomputed over 200 random partitions, and the distribution was summarised.

\subsection{Pre-specified Primary Hypothesis and Exploratory Family}
A hierarchical (gatekeeping) testing structure was adopted \cite{dmitrienko2011}. I-309/CCL1 was designated the single \emph{primary} predictive hypothesis \emph{a priori}, justified independently of the present cohort by Rangel Table~3 ($p<0.001$) \cite{rangel2025} and external chemokine evidence \cite{zhou2023,gupta2017}; it was tested at $\alpha=0.05$ on the formal partition. The remaining 17 inflammatory markers formed a \emph{secondary exploratory family} to which Benjamini-Hochberg false-discovery-rate (BH-FDR) control was applied \cite{benjamini1995}. Formally, given the ordered candidate $p$-values $p_{(1)}\le\cdots\le p_{(m)}$ with $m$ the family size, BH-FDR reports $q$-values $q_{(i)}=\min_{j\ge i}(m/j)\,p_{(j)}$ and controls the expected false-discovery proportion at level $\alpha$; candidates with $q\le0.05$ are declared discoveries. BH-FDR for the entire frozen 18-candidate family ($m=18$) is reported so the exploratory panel-wide result is visible alongside the primary one. The inflammatory family, correction method, primary hypothesis, and reporting partition were fixed before evaluation and were not adjusted to obtain significance; features with negative $\Delta$AUC are never interpreted as improvements.

\subsection{Calibration, Power, and Comparators}
Calibration was assessed on pooled OOF predictions using the Brier score, calibration-in-the-large (mean predicted risk minus prevalence), and the logistic calibration intercept and slope \cite{vancalster2019}. Because Naive Bayes posteriors can be overconfident, leakage-safe beta recalibration \cite{kull2017} was fitted on training-fold in-sample predictions and applied to the held-out fold for the baseline and baseline\,+\,I-309 models. From the observed effect and its test statistic, the sample size required for $80\%$ power was projected at the nominal $\alpha=0.05$ and at the family-adjusted threshold; this projection characterises screen power and was not used to rescue any result.

To check whether discrimination is lost through interpretability, the BNB/CNB was compared, on identical folds, against logistic regression, random forest, histogram gradient boosting, and a soft-score ensemble on the baseline\,+\,I-309 feature set (per-repeat paired DeLong versus the BNB) \cite{christodoulou2019}. As a sensitivity check, a fixed-partition screen was run in which the 18-candidate incremental screen was evaluated by every one of these model families, with BH-FDR applied within each model.

\section{Results}

\subsection{Baseline and the Incremental Screen}
\begin{figure}[t]
\centering
\includegraphics[width=0.9\linewidth]{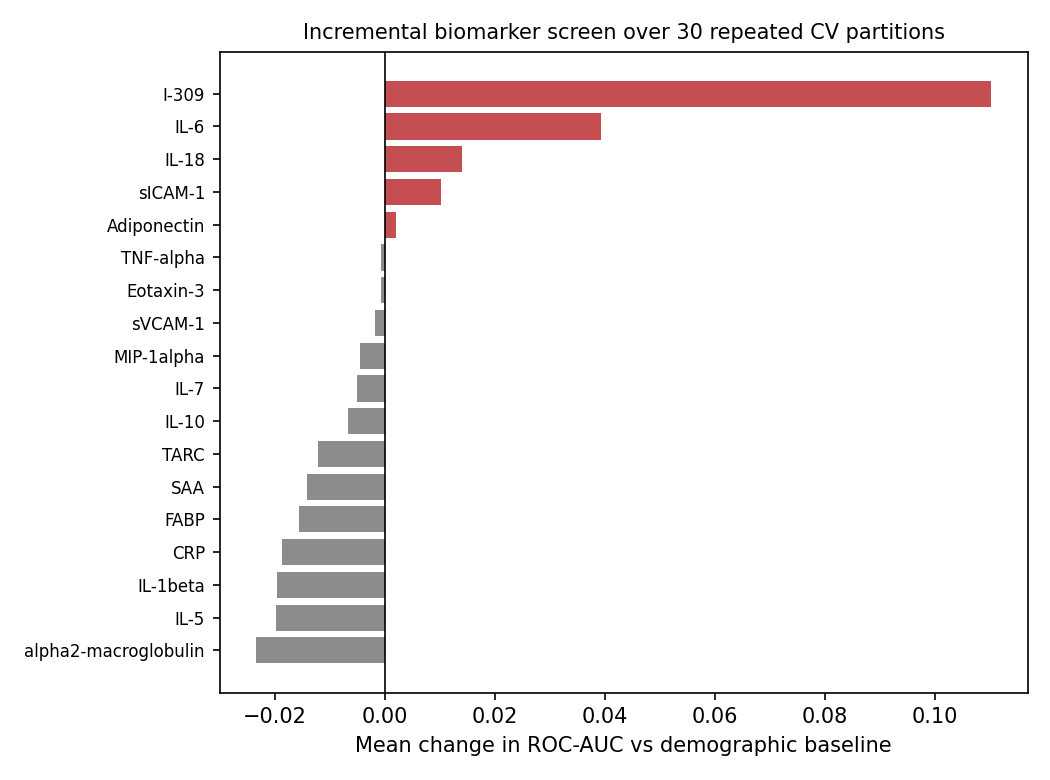}
\caption{Mean change in ROC-AUC versus the age + sex + education + income baseline. I-309 is the dominant positive incremental predictive feature; little or negative discrimination is added by most of the panel.}
\label{fig:delta}
\end{figure}

Repeated-CV ROC-AUC $0.630\pm0.017$ was achieved by the demographic baseline with income; $0.642\pm0.017$ was reached without income, but income was retained as part of the adjustment set. The pre-specified screen is shown in Table~\ref{tab:screen} and visualised in Fig.~\ref{fig:delta}. In predictive terms, the association findings did not transfer uniformly into useful features. I-309 was the only candidate with repeat-stable incremental discrimination: AUC $0.740\pm0.014$ was achieved by baseline\,+\,I-309, corresponding to mean $\Delta$AUC $+0.110$, median paired DeLong $p=0.001$, and nominal significance in $100\%$ of repeats. Only small deltas were produced by the remaining markers; IL-6 was the next largest at $\Delta$AUC $+0.039$, an order of magnitude weaker.

\subsection{Primary Hypothesis: I-309 Is Internally Stable}
On the frozen partition, paired DeLong $p=0.0018$ was obtained from the pre-specified primary test for I-309, below $\alpha=0.05$. This did not depend on a split: over 200 random partitions, median formal $p$ was $0.0011$ (interquartile range $0.0005$--$0.0019$), and the formal $p$ remained below $0.05$ across the resampled partitions. Because these partitions are resamples of one cohort rather than independent studies, this is read as internal validation: the pre-specified primary result, that I-309 adds discrimination beyond demographics, is internally stable.

\subsection{Exploratory Panel-Wide Screen}
Under BH-FDR for the 18-candidate family, $q=0.032$ was reached by I-309 on the frozen partition, i.e. $q<0.05$ was cleared, and no other candidate came close (next smallest $q=0.233$). However, this panel-wide result is boundary-sensitive: in the 200 partitions, median panel-wide $q$ for I-309 was $0.019$ (IQR $0.009$--$0.035$), and $q$ fell below $0.05$ in $85\%$ of partitions, not all. Panel-wide certification is therefore not claimed. The supported reading is that I-309 is significant as the pre-specified primary predictive hypothesis and usually survives panel-wide multiplicity, while no reliable incremental signal is shown by the rest of the inflammatory panel at this sample size.

\subsection{Interpretable Reference States}
Reference states for features are reported in Table~\ref{tab:ref}. The I-309 threshold was stable: the full-data serum concentration threshold was $5.34$ pg/mL, and the fold-level median and interquartile range were $5.34$ $[5.34, 5.34]$ pg/mL. Observed impairment prevalence was $0.42$ below that threshold and $0.87$ at or above it. The distinction between association and feature utility is illustrated by TNF-$\alpha$: a prevalence gap ($+0.22$) is shown, yet no repeated-CV discrimination over the baseline is added. Threshold shifts and incremental discrimination must therefore be reported.

\begin{table*}[!t]
\centering
\caption{Incremental predictive screen over the frozen 18-feature inflammatory discovery family. Mean values are over 30 repeated ten-fold CV partitions; the formal $p$ and BH $q$ columns use the fixed pre-specified partition. I-309 is the pre-specified primary hypothesis; the remaining rows are the secondary exploratory family.}
\label{tab:screen}
\scriptsize
\begin{tabular}{lcccccc}
\toprule
Candidate & AUC & $\Delta$AUC & Median $p$ & Repeats $p<0.05$ & Formal $p$ & BH $q$ \\
\midrule
\textbf{I-309} & \textbf{0.740$\pm$0.014} & \textbf{+0.110} & \textbf{$<$0.001} & \textbf{100} & \textbf{0.0018} & \textbf{0.032} \\
IL-6 & 0.669$\pm$0.016 & +0.039 & 0.097 & 10 & 0.039 & 0.233 \\
IL-18 & 0.644$\pm$0.017 & +0.014 & 0.349 & 0 & 0.206 & 0.529 \\
sICAM-1 & 0.640$\pm$0.018 & +0.010 & 0.459 & 0 & 0.284 & 0.535 \\
Adiponectin & 0.632$\pm$0.020 & +0.002 & 0.692 & 0 & 0.851 & 0.851 \\
TNF-$\alpha$ & 0.629$\pm$0.016 & $-$0.001 & 0.643 & 0 & 0.603 & 0.851 \\
Eotaxin-3 & 0.629$\pm$0.022 & $-$0.001 & 0.602 & 0 & 0.651 & 0.851 \\
sVCAM-1 & 0.628$\pm$0.017 & $-$0.002 & 0.688 & 0 & 0.778 & 0.851 \\
MIP-1$\alpha$ & 0.625$\pm$0.017 & $-$0.004 & 0.579 & 3 & 0.824 & 0.851 \\
IL-7 & 0.625$\pm$0.020 & $-$0.005 & 0.756 & 0 & 0.844 & 0.851 \\
IL-10 & 0.623$\pm$0.020 & $-$0.007 & 0.658 & 0 & 0.194 & 0.529 \\
TARC & 0.618$\pm$0.017 & $-$0.012 & 0.295 & 0 & 0.704 & 0.851 \\
SAA & 0.616$\pm$0.018 & $-$0.014 & 0.363 & 3 & 0.244 & 0.535 \\
FABP & 0.614$\pm$0.020 & $-$0.016 & 0.351 & 3 & 0.149 & 0.529 \\
CRP & 0.611$\pm$0.022 & $-$0.019 & 0.311 & 7 & 0.033 & 0.233 \\
IL-1$\beta$ & 0.610$\pm$0.020 & $-$0.020 & 0.140 & 30 & 0.087 & 0.390 \\
IL-5 & 0.610$\pm$0.020 & $-$0.020 & 0.144 & 23 & 0.369 & 0.604 \\
$\alpha_2$-macroglobulin & 0.606$\pm$0.019 & $-$0.023 & 0.070 & 47 & 0.297 & 0.535 \\
\bottomrule
\end{tabular}
\end{table*}

\begin{table}[t]
\centering
\caption{Interpretable reference states for selected leading inflammatory features. Thresholds are serum concentrations in pg/mL, shown as full-data or fold-median [IQR]; prevalences are observed, not adjusted.}
\label{tab:ref}
\scriptsize
\setlength{\tabcolsep}{2pt}
\begin{tabular*}{\columnwidth}{@{\extracolsep{\fill}}>{\raggedright\arraybackslash}p{0.20\columnwidth}>{\raggedright\arraybackslash}p{0.39\columnwidth}>{\centering\arraybackslash}p{0.17\columnwidth}>{\centering\arraybackslash}p{0.17\columnwidth}@{}}
\toprule
Feature & Threshold/state, pg/mL & Low/absent & High/present \\
 & & $n$, prev. & $n$, prev. \\
\midrule
I-309 & 5.34 [5.34, 5.34] & 100, 0.42 & 60, 0.87 \\
IL-6 & 0.781 [0.250, 0.781] & 29, 0.31 & 131, 0.65 \\
IL-18 & 345 [345, 345] & 146, 0.62 & 14, 0.21 \\
sICAM-1 & 279 [279, 279] & 49, 0.45 & 111, 0.65 \\
Adiponectin & $2.57\times10^4$ [$2.46\times10^4$, $2.57\times10^4$] & 93, 0.51 & 42, 0.71 \\
TNF-$\alpha$ & 2.25 [2.25, 2.39] & 85, 0.48 & 75, 0.71 \\
\bottomrule
\end{tabular*}
\end{table}

\subsection{Calibration}
Pooled-OOF calibration is reported in Table~\ref{tab:calib}. Raw BNB posteriors were overconfident, with calibration slopes below 1.0 (baseline $0.48$; baseline\,+\,I-309 $0.68$). Adding I-309 improved discrimination and the Brier score ($0.237\to0.205$), and moved the slope toward 1.0. Discrimination was unchanged by beta recalibration, and the Brier score was not improved; therefore, the posteriors should be read qualitatively. Rank discrimination is trustworthy, but absolute risk should not be taken as a deployable calculator.

\subsection{Power}
At the observed effect, the primary test is powered: roughly $n\approx133$ is sufficient for $80\%$ power at nominal $\alpha=0.05$, below the available $n=165$. Reliable panel-wide multiplicity is more demanding: roughly $n\approx249$ would be required for the family-adjusted threshold to be reached with $80\%$ power. This quantifies why the primary conclusion is stable while $q<0.05$ is cleared by the panel-wide screen $85\%$ of the time.

\subsection{Interpretability Costs No Discrimination}
On the feature set and folds, the interpretable BNB/CNB (AUC $0.740\pm0.014$) was statistically indistinguishable from logistic regression ($0.729$, paired $\Delta-0.011$), histogram gradient boosting ($0.723$, $\Delta-0.017$), and the soft ensemble ($0.743$, $\Delta+0.003$), and the random forest was exceeded ($0.709$, $\Delta-0.031$); the BNB was beaten by no comparator in more than $3\%$ of repeats (Table~\ref{tab:comp}). In the fixed-partition cross-model screen, I-309 was the only marker to survive within-model BH-FDR, and survival occurred under the interpretable model (q $=0.032$) and, more strongly, under the random forest (q $=0.005$); no other candidate survived under any model. Consistent with prior systematic evidence \cite{christodoulou2019}, added model complexity bought no discrimination for this parsimonious, auditable model, so the interpretable model is preferred.

\begin{table}[t]
\centering
\caption{Matched comparator sweep on baseline\,+\,I-309 (same folds, 30 repeats). $\Delta$AUC and $p$ are paired DeLong of each model versus the BNB/CNB.}
\label{tab:comp}
\begin{tabular}{lccc}
\toprule
Model & AUC & $\Delta$AUC vs BNB & Median $p$ \\
\midrule
BNB/CNB (proposed) & 0.740$\pm$0.014 & N/A & N/A \\
Logistic regression & 0.729$\pm$0.009 & $-$0.011 & 0.641 \\
Random forest & 0.709$\pm$0.015 & $-$0.031 & 0.232 \\
Hist.\ gradient boosting & 0.723$\pm$0.018 & $-$0.017 & 0.528 \\
Soft-score ensemble & 0.743$\pm$0.011 & +0.003 & 0.762 \\
\bottomrule
\end{tabular}
\end{table}

\begin{table}[t]
\centering
\caption{Pooled OOF calibration over 30 repeats. CITL is calibration-in-the-large (mean predicted risk minus prevalence).}
\label{tab:calib}
\begin{tabular}{lcccc}
\toprule
Model & AUC & Brier & Slope & CITL \\
\midrule
Baseline (raw) & 0.630 & 0.237 & 0.482 & +0.001 \\
Baseline (beta) & 0.626 & 0.236 & 0.518 & $-$0.002 \\
Baseline + I-309 (raw) & 0.740 & 0.205 & 0.683 & +0.004 \\
Baseline + I-309 (beta) & 0.737 & 0.206 & 0.675 & +0.001 \\
\bottomrule
\end{tabular}
\end{table}

\section{Discussion}

The association results from the study presented by Rangel et. al. \cite{rangel2025} were converted into a predictive question with an explicit, pre-specified structure, and the answer is selective: not every statistically associated biomarker becomes a useful predictive feature. The bar is higher than association alone, because a predictive feature must be learned on one subset of participants and evaluated on held-out participants. By that standard, I-309/CCL1 is the exception in this panel: it is the dominant marker in effect size and resampling stability, its pre-specified primary test gives $p=0.0018$, and its learned threshold ($5.34$ pg/mL, with impairment prevalence rising from $0.42$ below to $0.87$ at or above) is interpretable and selected. Panel-wide, the evidence is weaker: I-309 clears BH-FDR on the frozen partition and in most partitions, but the single-partition screen is near its power limit at $n=165$, so over-statement is avoided and no other marker shows reliable incremental value.

The pre-specified primary design is appropriate because it avoids dependence on the panel-wide correction. The I-309 hypothesis was privileged on external grounds, including Rangel et. al. association result and independent chemokine literature \cite{rangel2025,zhou2023,gupta2017}, rather than because that marker won the predictive screen; what those sources predicted is confirmed in a machine-learning setting. TNF-$\alpha$ provides the opposite lesson: although statistically relevant in the source analysis and descriptively shifted here, it did not improve repeated-CV discrimination beyond demographics.

The calibration analysis adds a caveat: raw Naive Bayes risks are overconfident and should be read as ranks, not deployable probabilities \cite{vancalster2019}. The results presented here show that statistical association does not always translate into predictive utility. I-309/CCL1 was the exception: both statistically associated and, here, an internally validated predictive feature. Reduced to a single, interpretable thresholded state, it is reusable in tabular-data prediction models of cognitive impairment. However, further external validation studies and evaluation of combinations with weaker features that may add complementary information are needed.

The modelling result also supports the BNB/CNB approach for translational screening on small clinical tabular data, the regime for which the framework was introduced \cite{thresholdbnb}. Deep learning is unrealistic at this sample size, and although flexible ensembles can match or exceed simple models they are not directly interpretable. The fitted model here is a small set of thresholds, class priors, and conditional probability tables rather than a black-box predictor; despite this simplicity, its discrimination was not statistically different from logistic regression, random forests, histogram gradient boosting, or the soft-score ensemble, and no more complex comparator provided a reliable advantage. This makes the BNB/CNB model a practical bridge between biomarker association analysis and clinically auditable prediction.

The present screen focused on independent feature testing. Each candidate was added to the baseline one at a time, and the BNB/CNB likelihood relies on conditional independence. This design is suited to finding strong single predictors because each marker's contribution can be isolated and audited. It does not rule out that combinations of weaker features could add incremental discrimination when modelled together. Testing such combinations, ideally with external validation and a pre-specified feature-selection strategy, is an appropriate direction for future work.

\section{Limitations}

The main limitations for this study is sample size. With $n=165$ and 18 feature columns, the screen is adequately powered for a single pre-specified primary predictive test but underpowered for simultaneous panel-wide certification, as shown by the partition-stability and power analyses. Evaluation remains same-cohort: although the screen is leakage-safe, an external cohort or nested selection would be required for unbiased performance estimation of any selected marker. Because Naive Bayes assumes conditional independence, interpretability is aided, but feature interactions may be missed when candidates are screened one at a time. The absolute posteriors are imperfectly calibrated and should not be used as a risk calculator, and numeric biomarker thresholds cannot be shown transportable from this cohort alone; external validation would be required to determine whether the same cut-points hold across assays, sites, and populations. Statistical inference has similar internal-validation limits. The repeated cross-validation partitions are resamples of a single cohort rather than independent studies, so the partition-stability figures describe internal stability under resampling and are not a substitute for external replication. DeLong's test was applied to out-of-fold cross-validated scores from models trained on overlapping folds; its variance assumptions are therefore only approximately met, and the reported $p$-values should be read as internal-validation summaries rather than exact inferential guarantees. Finally, the PARI-HD dataset is confidential and cannot be redistributed, though the full analysis code is released so the workflow is auditable \cite{garcia2026repo}.

\section{Ethics and Data-Use Statement}

This secondary analysis of IRB-approved PARI-HD data followed the source-study framework, including approval SDMdeDEI-CH-248-12 and the Declaration of Helsinki \cite{rangel2025}. No new participants were recruited, and no new consent was required. 

\section{Conclusion}

With a leakage-safe, interpretable BNB/CNB workflow, biomarkers that had been previously reported as statistically associated with cognitive impairment were re-examined as machine learning prediction features rather than assumed to be predictive. Only one survived this stricter test: I-309/CCL1 added stable, auditable discrimination beyond the demographic baseline, with a pre-specified primary result of $p=0.0018$ that was internally stable across resampling partitions; it cleared panel-wide false-discovery control on the frozen partition and in most, though not all, partitions, while no other marker cleared it. The transferable contribution is the procedure itself: a reproducible, interpretable way to decide which statistical associations actually earn a place in a predictive model. Applied to this cohort, it leaves a single concrete candidate, an I-309/CCL1 threshold near $5.34$ pg/mL that is directly reusable as a feature in downstream tabular data models, and defines the decisive next step of external validation in an independent cohort.

\balance

\end{document}